\documentclass{isprs} % isprs class modified 23-04-2019 (Dennis Wittich)
\usepackage{graphicx}
\usepackage{array}
\usepackage{booktabs} 
\usepackage{multirow}
\usepackage{amsmath}
\usepackage{amssymb}
\usepackage{subfigure}
\usepackage{setspace}
\usepackage{geometry} % added 27-02-2014 Markus Englich
\usepackage{epstopdf}
\usepackage[labelsep=period]{caption}  % added 14-04-2016 Markus Englich - Recommendation by Sebastian Brocks
\usepackage[british]{babel} 
\usepackage[hang]{footmisc}
\usepackage{xcolor}
\usepackage{enumitem}
\usepackage[justification=justified]{caption}

\begin{document}

\title{A Comparison of Multi-View Stereo Methods for Photogrammetric 3D Reconstruction: From Traditional to Learning-Based Approaches    
}

\author{
Y. Li\textsuperscript{1}, G. Vosselman\textsuperscript{1}, F. Nex\textsuperscript{1} }

% KAO: Remove extra newline
\address{
	\textsuperscript{1 }Faculty of Geo-Information Science and Earth Observation (ITC), University of Twente, Enschede, The Netherlands \\
	 (yawen.li, george.vosselman, f.nex)@utwente.nl\\
}

% If the corresponding author is NOT the final author, always add a % space before the subsequent comma, i.e.
% first author name\textsuperscript{a,}\thanks{Corresponding author} , % second author name \textsuperscript{b}, etc.
% thanks to Niclas Borlin 05-05-2016
% information on the corresponding author should not be used any longer and has been commented out
% C. Heipke, Jan 03,2024

% the use of the information of commissions and working groups should not be used any longer and has been commented out
% C. Heipke, Sept. 20,2022
%\commission{XX, }{YY} %This field is optional. If filled, XX and YY should be replaced by adequate numbers. See https://www2.isprs.org/commissions/
%\workinggroup{XX/YY} %This field is optional.
%\icwg{}   %This field is optional.

\abstract{

Photogrammetric 3D reconstruction has long relied on traditional Structure-from-Motion (SfM) and Multi-View Stereo (MVS) methods, which provide high accuracy but face challenges in speed and scalability. Recently, learning-based MVS methods have emerged, aiming for faster and more efficient reconstruction. This work presents a comparative evaluation between a representative traditional MVS pipeline (COLMAP) and state-of-the-art learning-based approaches, including geometry-guided methods (MVSNet, PatchmatchNet, MVSAnywhere, MVSFormer++) and end-to-end frameworks (Stereo4D, FoundationStereo, DUSt3R, MASt3R, Fast3R, VGGT). Two experiments were conducted on different aerial scenarios. The first experiment used the MARS-LVIG dataset, where ground-truth 3D reconstruction was provided by LiDAR point clouds. The second experiment used a public scene from the Pix4D official website, with ground truth generated by Pix4Dmapper. We evaluated accuracy, coverage, and runtime across all methods. Experimental results show that although COLMAP can provide reliable and geometrically consistent reconstruction results, it requires more computation time. In cases where traditional methods fail in image registration, learning-based approaches exhibit stronger feature-matching capability and greater robustness. Geometry-guided methods usually require careful dataset preparation and often depend on camera pose or depth priors generated by COLMAP. End-to-end methods such as DUSt3R and VGGT achieve competitive accuracy and reasonable coverage while offering substantially faster reconstruction. However, they exhibit relatively large residuals in 3D reconstruction, particularly in challenging scenarios.}


\maketitle

\section{Introduction}
\sloppy

With the rapid growth of vision-based systems, the ability to reconstruct 3D environments from images has become increasingly important. Accurate and efficient 3D reconstruction is essential for advanced tasks such as scene understanding, embodied intelligence, and large-scale digital twins \cite{jiang20243d}. Among various application domains, photogrammetric 3D reconstruction represents one of the most widely adopted scenarios \cite{jiang2021unmanned}. It derives precise geometric and spatial information from overlapping images, enabling large-scale mapping and metric measurements across aerial scenarios.

3D reconstruction can be generally divided into single-view and multi-view methods \cite{facil2017single}. Due to its simplicity and flexibility, single-view 3D reconstruction has attracted growing attention in recent years. However, such approaches heavily rely on learned priors and lack explicit geometric constraints, which limits their accuracy, generalization, and metric reliability in real-world scenarios \cite{kato2019learning,sun2021neuralrecon}. In parallel, neural rendering methods such as NeRF \cite{mildenhall2021nerf} and 3D Gaussian Splatting (3DGS) \cite{kerbl20233d} have demonstrated impressive performance in novel view synthesis.  Their primary objective is to generate visually plausible images from unseen viewpoints, focusing on visual fidelity rather than geometric accuracy \cite{gupta2024comparison}. In contrast, Multi-View Stereo (MVS) based methods are explicitly designed to recover metric 3D structure from images. Therefore, this paper focuses on traditional and learning-based MVS frameworks, which are more suitable for high-accuracy aerial photogrammetric reconstruction.

Compared with single-view methods, multi-view reconstruction leverages geometric consistency across multiple viewpoints, which reduces shape ambiguity and improves robustness. These advantages make multi-view reconstruction approaches particularly suitable for high-precision and large-scale photogrammetric applications. Traditional multi-view reconstruction approaches, such as Structure-from-Motion (SfM) and MVS, have made significant progress in the past decades. Representative open-source systems such as COLMAP \cite{schonberger2016pixelwise,schonberger2016structure}, OpenMVS \cite{cernea2020openmvs}, and OpenMVG \cite{moulon2016openmvg}, as well as widely used commercial software such as Pix4Dmapper \cite{vallet2012photogrammetric} and Agisoft Metashape \cite{verhoeven2011taking}, have become mainstream tools for photogrammetric 3D reconstruction \cite{liu2023deep}. Using only images as input, these systems implement modular pipelines for feature extraction, feature matching, sparse reconstruction, and dense stereo fusion, enabling camera pose estimation and dense 3D reconstruction \cite{zhao2021rtsfm}. Nevertheless, they still face notable bottlenecks, including limited real-time performance and reliance on complex multi-stage pipelines.

\begin{figure*}[ht!]
\begin{center}
    \includegraphics[width=6.5in]{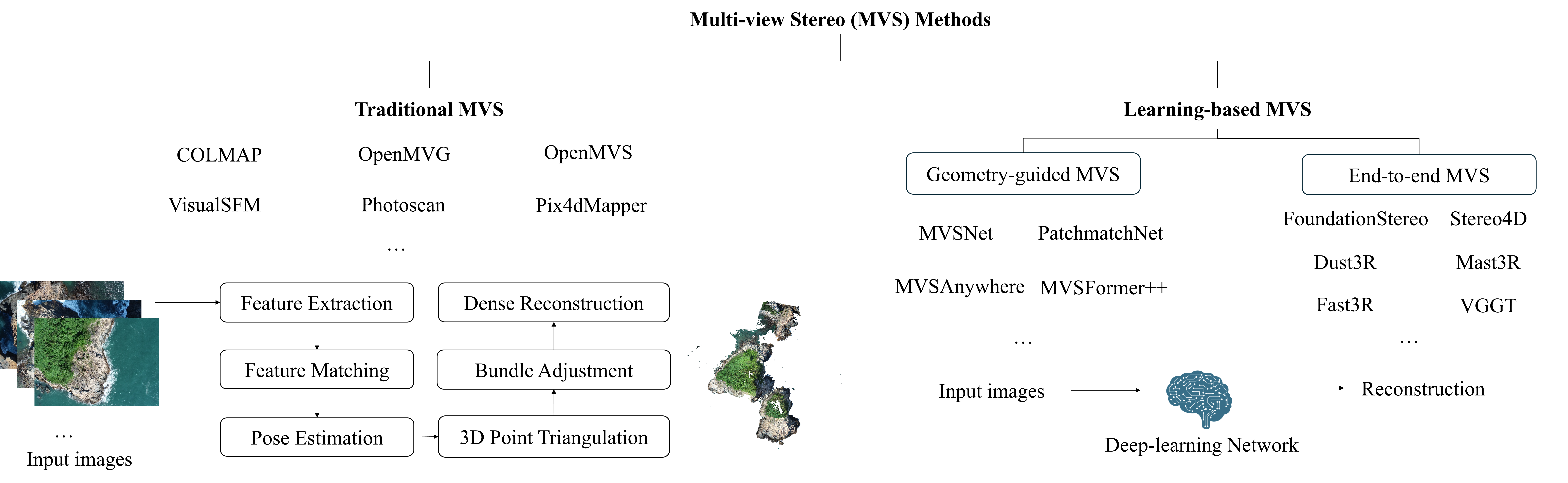}  % Adjusts the figure to the width of the page
    \captionsetup{justification=centering, singlelinecheck=false}
    \centering
    \caption{Multi-View Stereo Methods.}
    \label{MVS}
\end{center}
\end{figure*}

With the rapid development of large-scale models, deep learning–based methods have emerged as promising alternatives for 3D reconstruction. These approaches eliminate the reliance on hand-crafted components of classical pipelines by directly estimating the 3D structure from images. Existing learning-based frameworks can be broadly categorized into geometry-guided MVS methods and end-to-end reconstruction methods, and they support both two-view and multi-view reconstruction settings. Two-view methods reconstruct 3D structures from image pairs. They are suitable for lightweight and rapid applications but offer limited geometric constraints. Multi-view methods exploit geometric consistency across multiple viewpoints to achieve more accurate and complete scene reconstruction.

Geometry-guided methods, such as MVSNet \cite{yao2018mvsnet}, CasMVSNet \cite{gu2020cascade}, PatchMatchNet \cite{wang2021patchmatchnet}, MVSAnywhere \cite{izquierdo2025mvsanywhere}, and MVSFormer++ \cite{cao2024mvsformer++} integrate traditional multi-view geometry with learned cost volumes to predict depth maps efficiently. These methods focus on densification, relying on images and externally estimated camera poses as input.

In contrast, end-to-end models, including FoundationStereo \cite{wen2025foundationstereo}, Stereo4D \cite{jin2024stereo4d}, DUSt3R \cite{wang2024DUSt3R}, MASt3R \cite{leroy2024grounding}, VGGT \cite{wang2025vggt}, and Fast3R \cite{yang2025fast3r}, leverage large-scale vision models to infer 3D structure directly from image collections without explicit geometric modeling. These approaches demonstrate strong capabilities in terms of speed, robustness, and generalization, showing great potential for scalable and real-time photogrammetric reconstruction.

Despite the rapid progress of learning-based MVS methods, few studies \cite{hermann2024depth} have systematically compared them with traditional frameworks in the field of photogrammetry. This work bridges that gap by evaluating learning-based approaches against a representative traditional approach. For experimental evaluation, we use two aerial datasets: the MARS-LVIG dataset with LiDAR-based ground truth \cite{li2024mars}, and a public Pix4D dataset with reference point clouds generated by Pix4Dmapper \cite{pix4d_example_dataset}. Quantitative and qualitative assessments are performed in CloudCompare \cite{girardeau2016cloudcompare}, reporting metrics such as processing time, accuracy, and coverage. Our goal is to provide insights into their strengths and weaknesses and offer useful references for downstream 3D reconstruction tasks.

The remainder of this paper is organized as follows. Section II provides an overview of the MVS methods, including traditional and learning-based MVS methods. Section III discusses the datasets, experimental results, and analysis. Finally, the conclusions and future work are summarized in the last section.

\section{Related Works}
In this section, we review the traditional and learning-based reconstruction frameworks. Fig. \ref{MVS} shows the overall 3D reconstruction pipeline. The process starts from a set of aerial images captured from a UAV platform. These aerial images are subsequently processed through two distinct methodological branches: (1) traditional MVS, which is based on classical pipelines, and (2) learning-based MVS, which leverages recent advances in deep learning.

\subsection{Traditional MVS Methods}

Traditional MVS frameworks take only images as input and typically consist of several well-defined stages, including feature extraction and matching, camera pose estimation, 3D point triangulation, bundle adjustment, and dense reconstruction.

COLMAP \cite{schonberger2016pixelwise,schonberger2016structure} is one of the most representative open source systems. It used incremental SfM combined with global BA optimization and generated high-quality dense point clouds. OpenMVG primarily focused on extracting dense feature matches from input images and used these matches to estimate camera pose and reconstruct sparse point clouds, providing a complete SfM framework \cite{moulon2016openmvg}. OpenMVS focused on dense point cloud processing and surface reconstruction, generating high-quality 3D models through multi-view depth fusion and texture mapping \cite{cernea2020openmvs}. It can also directly use OpenMVG output as input, achieving more refined reconstruction results.

In the field of aerial photogrammetric mapping, several mature software platforms have been widely adopted for practical applications. Agisoft Metashape \cite{verhoeven2011taking} and Pix4Dmapper \cite{vallet2012photogrammetric} are two widely used commercial software packages that provide highly automated and robust photogrammetric workflows. They integrated aerial triangulation, dense image matching, surface modeling, and texture mapping within a unified framework.

Traditional MVS pipelines rely on accurate image orientation, which necessitates precise registration and parameter tuning. Any failure during this stage can propagate through subsequent reconstruction steps, leading to degraded results. Moreover, they often suffer from limited real-time capabilities, which can be computationally expensive and take hours on large-scale datasets. They are also sensitive to textureless or repetitive regions, illumination changes, and image quality, which often lead to visible artifacts and distorted reconstructions. Such limitations are further amplified in aerial photogrammetry, where high-resolution imagery, wide-area coverage, and stringent accuracy requirements impose additional challenges on both computational efficiency and reconstruction quality.

\begin{table*}[t!]
\begin{tabular}{l c p{8cm} l}
\hline
\multicolumn{1}{c}{{\color[HTML]{000000} Method}} & \multicolumn{1}{c}{{\color[HTML]{000000} Number of Views}} & \multicolumn{1}{c}{{\color[HTML]{000000} Characteristic}} & \multicolumn{1}{c}{{\color[HTML]{000000} Category}}           \\ \hline
 COLMAP                     &  N           & Classical SfM+MVS pipeline with explicit geometry           & Traditional                           \\ \hline
PatchmatchNet             & N                                   &   Learned PatchMatch for cost volume–based depth regression                             & Geomery-Guided                        \\ \hline
 MVSFormer++               & N                                   &  Transformer-based cost aggregation with explicit geometry                                  & Geomery-Guided                         \\ \hline
MVSAnywhere                                       & N                                                          &   Cost Volume Patchifier with view-independent and scale-independent geometry                                                        & Geomery-Guided \\ \hline
FoundationStereo                                 & 2                                                          &  Zero-shot generalization                                                        & End-to-end                                                \\ \hline
Stereo4D                                          & 2                                                          &  Monocular priors + 4D cost volume with temporal consistency                                                      & End-to-end                                                \\ \hline
DUSt3R                                            & 2                                                          &   ViT-based dense correspondence with pointmap regression                                                        & End-to-end                                                    \\ \hline
MASt3R                                            & 2                                                          &  Extension of DUSt3R with enhanced feature extraction and fast reciprocal matching                                                         & End-to-end                                                    \\ \hline
Fast3R                                            & N                                                          &   Efficient and scalable end-to-end matching via positional interpolation                                                       & End-to-end                                                    \\ \hline
VGGT                                              & N                                                          &   Large Transformer with global and frame attention                                                        & End-to-end                                                    \\ \hline
\end{tabular}
\caption{Overview and categorization of representative MVS methods considered in this study}
\label{tab:selection}
\end{table*}

\subsection{Learning-based MVS Methods}

Recently, several learning-based approaches have been developed for generating depth maps and reconstructing 3D models. These methods can be broadly categorized into geometry-guided MVS methods and end-to-end MVS methods. As summarized in Table 1, the number of views indicates
whether the method supports two-view (2) or multi-view (N)
reconstruction, which directly impacts its scalability to largescale
mapping scenarios. The characteristic briefly describes
the core mechanism of each method, highlighting whether it relies
on explicit geometric modeling, geometry-guided learning,
or end-to-end reasoning. The category column classifies methods
into three main classes: traditional, geometry-guided, and
end-to-end.

\subsubsection{Geometry-guided MVS methods}

\noindent

By integrating multi-view geometry into the learning framework, these methods take images together with externally estimated poses as inputs, and construct cost volumes from camera parameters to enable accurate dense depth estimation.
    
MVSNet \cite{yao2018mvsnet} constructed a cost volume based on the reference image and employed CNNs to learn 3D regularization. To improve memory consumption, R-MVSNet \cite{yao2019recurrent} replaced MVSNet's depth prediction with a recurrent network to make improvements. Subsequently, many approaches began to explore how to reduce computing resources and improve operating efficiency. Cas-MVSNet \cite{gu2020cascade} put forward a cascade cost volume and VA-MVSNet \cite{yi2020pyramid} introduced a self-adjusting view aggregation to determine the contribution of each input image on a voxel-by-voxel basis. PatchmatchNet \cite{wang2021patchmatchnet} adapted patchmatch-based propagation to replace heavy 3D convolutions, enabling iterative depth estimation in a coarse-to-fine manner.

More recent transformer-based or hybrid methods, such as MVSFormer, MVSFormer++, and MVSAnywhere, leveraged Vision Tranformers (ViTs) to capture long-range dependencies and enhance cross-view reasoning. MVSFormer \cite{caomvsformer} enhanced MVS depth estimation by fusing pretrained ViT features and employing multi-scale training with hybrid classification–regression for improved robustness. MVSFormer++ \cite{cao2024mvsformer++} further integrated DINOv2 \cite{oquab2023dinov2} features with a proposed side view attention, achieving higher accuracy and generalization in challenging multi-view scenarios. MVSAnywhere \cite{izquierdo2025mvsanywhere} was a transformer-based MVS framework that introduced a cost volume patchifier, enabling tokenization of cost volumes while fusing single-view ViT features. It employed a view-independent and scale-independent cost volume construction mechanism, achieving state-of-the-art depth accuracy and 3D consistency across arbitrary numbers of input views.

The aforementioned methods are mostly evaluated on indoor or simulated datasets. These scenes typically have decent lighting, limited scale, and clear geometric structure, which simplifies the reconstruction problem. The generalization ability of these methods to large-scale, high-resolution outdoor scenes remains underexplored.

\begin{table*}[t!]
\centering
\begin{tabular}{
  >{\centering\arraybackslash}m{2cm}
  >{\centering\arraybackslash}m{2cm}
  >{\centering\arraybackslash}m{2cm}
  >{\centering\arraybackslash}m{2cm}
  >{\centering\arraybackslash}m{2cm}
  >{\centering\arraybackslash}m{3cm}}
\hline
\multicolumn{1}{c}{{\color[HTML]{000000} Dataset}} & 
\multicolumn{1}{c}{{\color[HTML]{000000} Scene}} & 
\multicolumn{1}{c}{{\color[HTML]{000000} Total images}} & 
\multicolumn{1}{c}{{\color[HTML]{000000} Groundtruth}} &
\multicolumn{1}{c}{{\color[HTML]{000000} Image Resolution}} &
\multicolumn{1}{c}{{\color[HTML]{000000} Display}} \\ 
\hline
{\color[HTML]{000000} MARS-LVIG} & 
{\color[HTML]{000000} Island} & 
{\color[HTML]{000000} 149} & 
{\color[HTML]{000000} LiDAR} &
{\color[HTML]{000000} 5472*3648} &
\includegraphics[width=2cm]{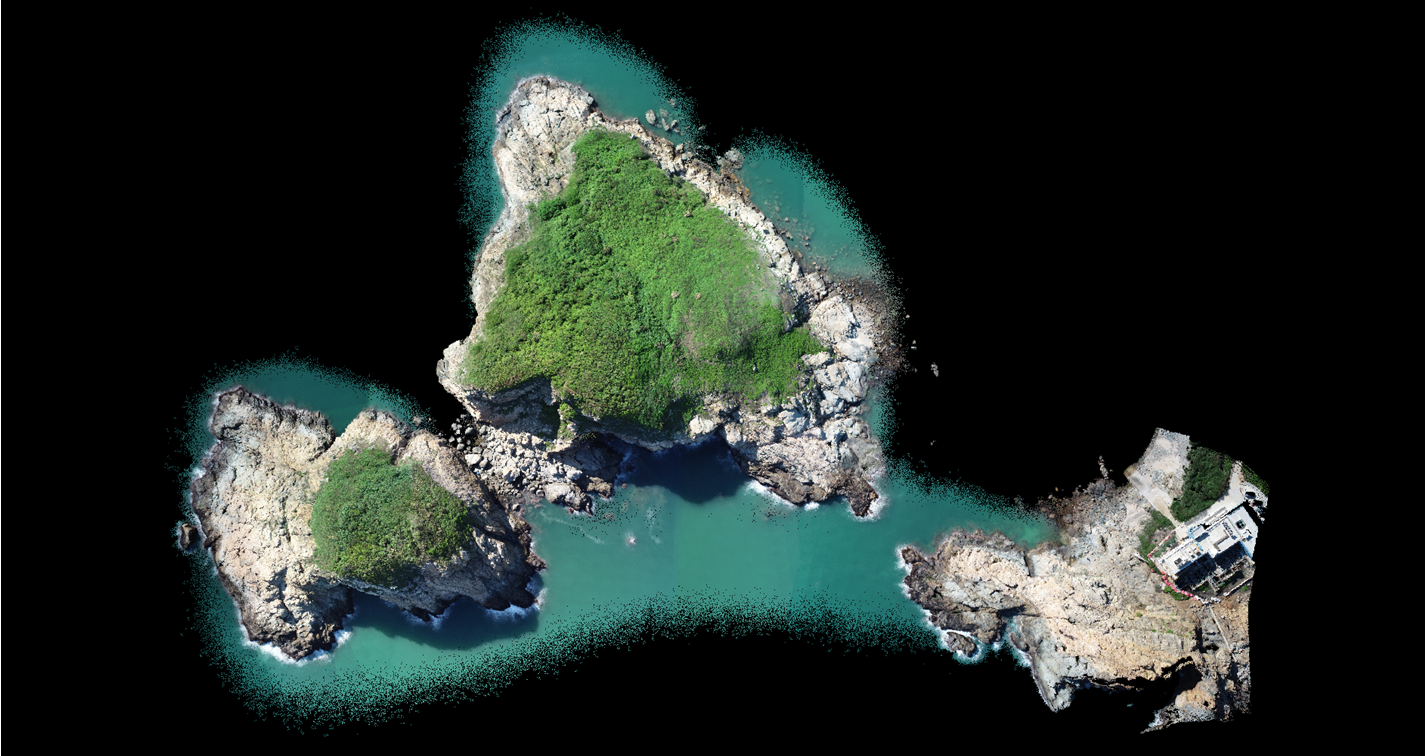} \\ 
\hline
{\color[HTML]{000000} Pix4D Example} & 
{\color[HTML]{000000} Urban area} & 
{\color[HTML]{000000} 100} & 
{\color[HTML]{000000} Pix4Dmapper} &
{\color[HTML]{000000} 6000*4000} &
\includegraphics[width=2cm]{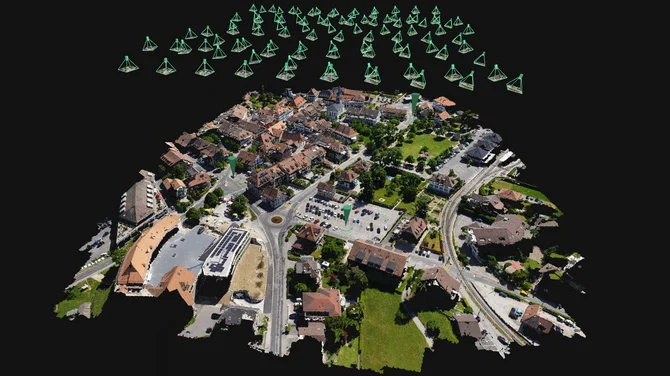} \\ 
\hline
\end{tabular}
\caption{Datasets used for our tests.}
\label{tab:dataset_display}
\end{table*}

\subsubsection{End-to-end MVS methods} 

\noindent

End-to-end MVS methods integrate correspondence estimation, pose recovery, and 3D structure generation into a unified learning pipeline using only images. By leveraging large-model architectures, these approaches are more flexible and directly produce dense and coherent 3D reconstructions.

Two-view methods such as Stereo4D and FoundationStereo focus on lightweight and efficient stereo inference, relying on rectified image pairs as input. Stereo4D \cite{jin2024stereo4d} focused on real-time stereo depth estimation for dynamic 3D scenes by jointly modeling spatial and temporal cues, enabling fast and robust 3D reconstruction under motion. FoundationStereo \cite{wen2025foundationstereo} leveraged large foundation models to achieve accurate and robust stereo matching with minimal task-specific training, providing strong generalization across diverse scenes.
    
DUSt3R \cite{wang2024DUSt3R} was a milestone in end-to-end 3D reconstruction, enabling direct and efficient point cloud generation from image pairs. It adopted a ViT-based architecture that was pre-trained for cross-view correspondence reasoning. Two input images were first fed into a shared encoder, and their features were then processed by a transformer-based decoder using cross-attention to establish dense correspondences. At the end of the decoder, two independent prediction heads output point maps for both views, with the second point map aligned to the coordinate frame of the first view. MASt3R \cite{leroy2024grounding} built upon DUSt3R by introducing a new feature extraction head that produced dense local features and employed a matching loss to enhance correspondence accuracy while maintaining robustness under challenging viewpoint changes. DUSt3R and MASt3R were limited to pairwise reconstruction, requiring additional fusion and geometric post-processing, and became inefficient when extended to multi-view settings. 

To overcome these scalability limitations, Fast3R \cite{yang2025fast3r} leveraged positional interpolation to train on a small number of views and generalized to large-scale multi-view inference. It integrated FlashAttention with parallel training and combined local point maps with global alignment to improve efficiency and consistency. VGGT \cite{wang2025vggt} employed a large-scale transformer architecture with a global attention and frame attention mechanism. In addition, a camera-head module was introduced to estimate camera parameters and 3D structure simultaneously.

\section{Experiments} 
\subsection{Method Selection}

To ensure a fair and representative comparison, this study selects a set of representative MVS methods, including COLMAP, PatchMatchNet, MVSFormer++, MVSAnywhere, FoundationStereo, Stereo4D, DUSt3R, MASt3R, VGGT, and Fast3R. In this study, we use the publicly available pretrained
models without any fine-tuning.

\subsection{Dataset}
To evaluate the performance of different reconstruction paradigms, we use two datasets shown in Table \ref{tab:dataset_display}. The MARS-LVIG datasets are UAV-based aerial photogrammetry datasets collected in complex environments \cite{li2024mars}. These datasets are well-suited for testing reconstruction robustness and scalability due to their high-resolution imagery, large-scale coverage, and complex structures. A challenging island scene is used in this study, with image resolutions of 5472 × 3648. The ground-truth reference is derived from high-precision LiDAR point clouds.

In addition, we include a publicly available Pix4Dmapper official dataset, consisting of 100 aerial frames \cite{pix4d_example_dataset}. This dataset provides well-structured image acquisition with sufficient overlap, enabling stable and reliable reconstruction. A typical urban scene is used for evaluation, with image resolutions of 6000 × 4000. The ground-truth point clouds are generated using Pix4Dmapper.

For geometry-guided methods, the inputs are derived from COLMAP-generated results, including camera poses, sparse reconstruction, undistorted images, and view selection files. To align the input conditions among different methods, all input images for DUSt3R/MASt3R/Fast3R/VGGT in our experiments were undistorted beforehand. For FoundationStereo and Stereo4D, the input image pairs were rectified beforehand to meet their stereo inference requirements.

DUSt3R, MASt3R, and Fast3R are limited to a maximum input size of 512 pixels, and VGGT requires 518 pixels. To ensure a fair and systematic comparison, all images were rescaled to 512 pixels to the maximum dimension. These rescaled images were then used as input for the rest methods \cite{wu2025evaluation}.

We employed a progressive subset selection strategy, sampling images in increasing numbers (2, 10, 20, 50, and 100 frames). It allows us to evaluate how each method scales with the number of input views and analyze their reconstruction accuracy, completeness, and runtime under different input conditions. For the subsets with 2, 10, and 20 images, the reference point clouds were cropped to match the corresponding coverage area.

\subsection{Evaluation Indicators}
We conduct a comprehensive and systematic evaluation of representative methods across all key aspects. Specifically, we assess their performance in terms of processing time, reconstruction accuracy, and coverage, ensuring a fair and consistent comparison under the same experimental conditions \cite{wu2025evaluation}.

Processing time (seconds [s]) reflects the overall computational efficiency of each method. It provides a practical measure of algorithm scalability and suitability for real-time and large-scale applications. The reported runtime for each image setting is averaged over three runs.

Accuracy quantifies the geometric closeness between the reconstructed point cloud and the reference ground truth. In this study, it is assessed using two commonly used indicators: Root Mean Square (RMS), which measures the overall deviation between corresponding points, and Mean Distance (MD), which reflects the average alignment accuracy across the entire point cloud. The island aerial images (5472 × 3648 pixels) have an average ground sampling distance (GSD) of 2.5 cm/pixel. After downsampling to 512 × 341 pixels, the effective GSD becomes approximately 0.267 m/pixel. The urban aerial images (6000 × 4000 pixels) have an average ground GSD of 2.41 cm/pixel. After downsampling to 512 × 341 pixels, the effective GSD becomes approximately 0.282 m/pixel. All residual errors (RMS and MD) are therefore reported in normalized units.

Coverage measures the proportion (\%) of the reference surface that is successfully reconstructed. It is defined as the percentage of ground truth points whose distances to the reconstructed surface are within a predefined tolerance. In our experiments, a threshold of 1 meter is applied.

All metrics were computed using CloudCompare \cite{girardeau2016cloudcompare}, ensuring a unified and objective evaluation framework across different methods.

\begin{figure*}[t]
\begin{center}
    \includegraphics[width=5.8in]{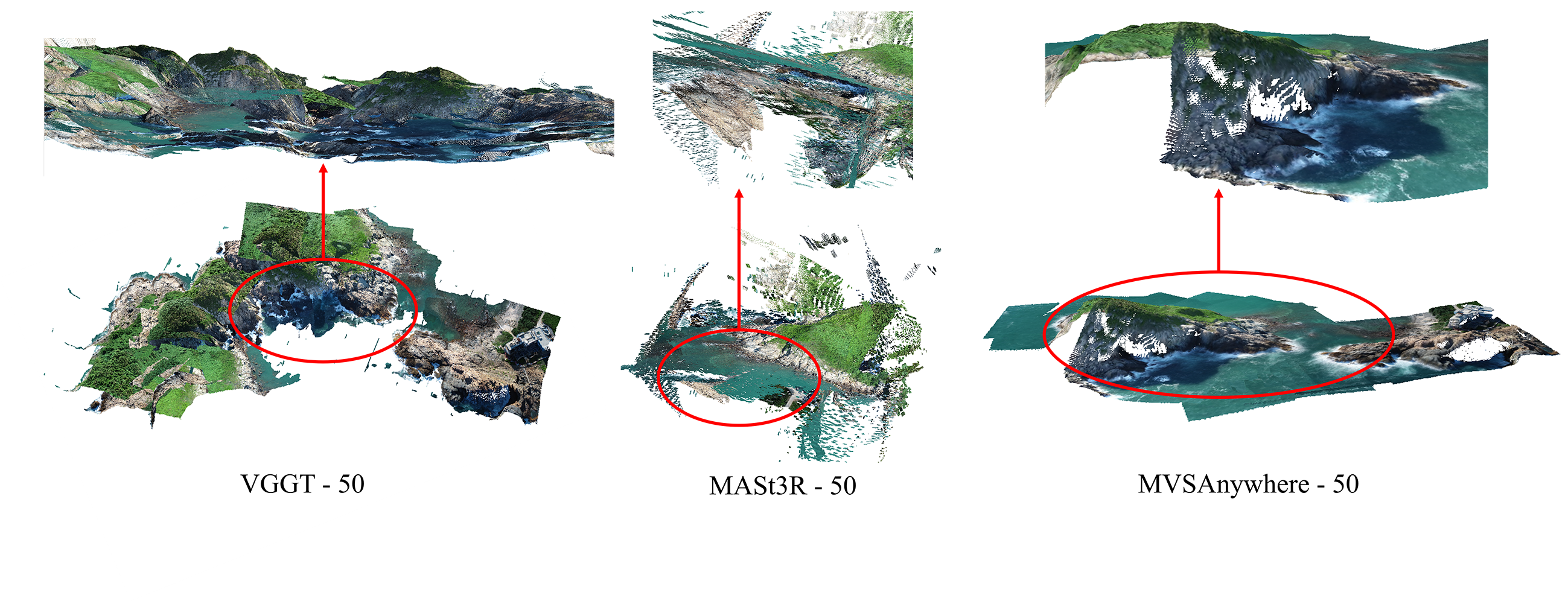}  % Adjusts the figure to the width of the page
    \captionsetup{justification=centering, singlelinecheck=false}
    \centering
    \caption{Failed results on Island. The notation “Method–N” (e.g., MASt3R-50, VGGT-50) indicates the reconstruction result obtained from different image samples (N). The red circle marks a local area with discontinuities and layering effects in the island reconstructed point cloud. The zoomed-in view above highlights these artifacts.}
    \label{failisland}
\end{center}
\end{figure*}

\begin{figure*}[t]
\begin{center}
    \includegraphics[width=5.8in]{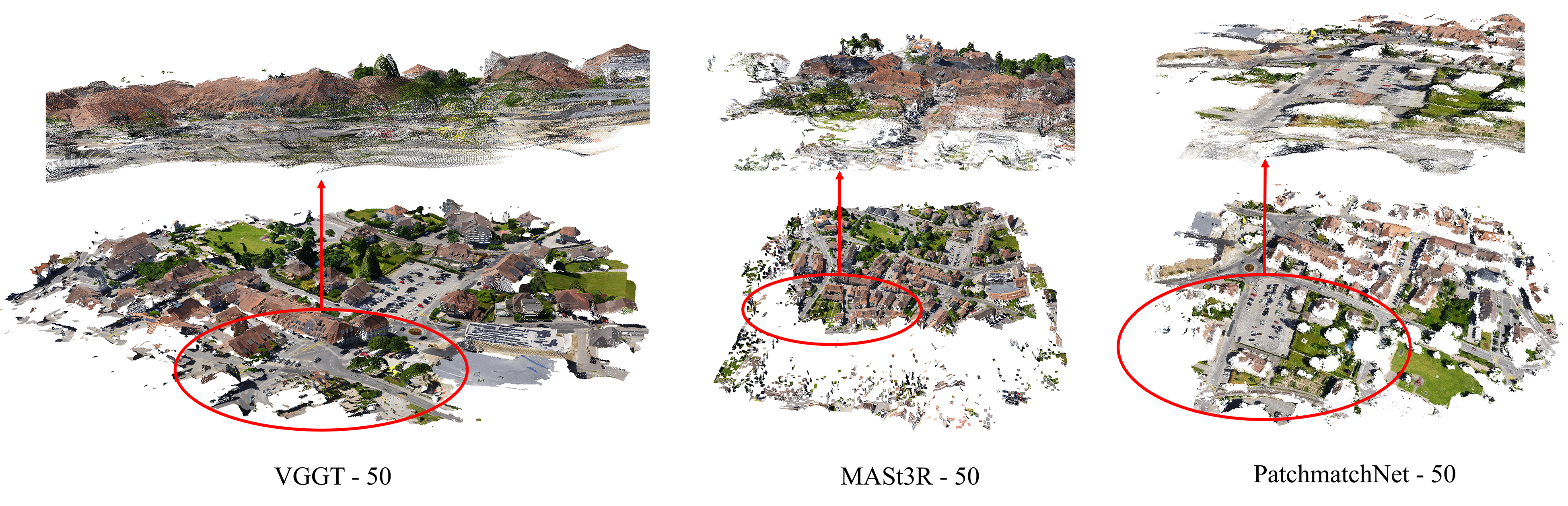}  % Adjusts the figure to the width of the page
    \captionsetup{justification=centering, singlelinecheck=false}
    \centering
    \caption{Failed results on Urban. The notation “Method–N” (e.g., MASt3R-50, VGGT-50) indicates the reconstruction result obtained from different image samples (N). The red circle marks a local area with discontinuities and layering effects in the urban reconstructed point cloud. The zoomed-in view above highlights these artifacts.}
    \label{failurban}
\end{center}
\end{figure*}

\subsection{Results}
The experimental evaluations were conducted using a single NVIDIA A40 GPU. These evaluation metrics are thoroughly analyzed in our experiments, providing quantitative insights among different approaches. In all tables, the numerically best performance for each metric is shown in bold, and the second-best performance is shown underlined.

\begin{table}[t]
\centering
\footnotesize
\renewcommand{\arraystretch}{1.2}
\begin{tabular}{
  >{\raggedright\arraybackslash}m{1.7cm}  
  >{\centering\arraybackslash}m{0.6cm}     
  >{\centering\arraybackslash}m{0.6cm}     
  >{\centering\arraybackslash}m{0.7cm}     
  >{\centering\arraybackslash}m{0.7cm}     
  >{\centering\arraybackslash}m{0.7cm}      
}
\toprule
\textbf{Methods} & \textbf{2} & \textbf{10} & \textbf{20} & \textbf{50} & \textbf{100} \\
\midrule
DUSt3R             & 3.15 & 35.91 & 120.16 &       &        \\
MASt3R             & 8.60 & 39.49 &  68.63 & 1075.49 &    \\
VGGT               & 1.52 &  \underline{6.10} &  \underline{16.59} &  \underline{28.36} & \underline{60.79}   \\
Fast3R             & \textbf{0.41} &  \textbf{0.82} &  \textbf{1.70} &  \textbf{5.57} &   \textbf{16.47}   \\
MVSAnywhere        &   N/A &   N/A  & N/A & 42.07 & 100.28   \\
MVSFormer++        & 15.34 &   64.99  &  79.29 &  80.42 & 304.28  \\
PatchmatchNet      & 5.67 &   15.51  & 23.76 & 34.45 & 238.10  \\
FoundationStereo   & 2.82 &     &     &      &       \\
Stereo4D           &  \underline{0.63} &     &      &      &        \\
\midrule
COLMAP           &  21.18 &   86.92  &   183.24   &   343.85   &   1326.38     \\
\bottomrule
\end{tabular}
\caption{Runtime (s) of different methods on Island across all subsets with varying numbers of input images (2, 10, 20, 50, and 100).  “N/A” indicates no result is available.}
\label{tab:runtime_island}
\end{table}

\subsubsection{Processing time}

\noindent

\begin{table}[t]
\centering
\footnotesize
\renewcommand{\arraystretch}{1.2}
\begin{tabular}{
  >{\raggedright\arraybackslash}m{1.7cm}  
  >{\centering\arraybackslash}m{0.6cm}     
  >{\centering\arraybackslash}m{0.6cm}     
  >{\centering\arraybackslash}m{0.7cm}     
  >{\centering\arraybackslash}m{0.7cm}     
  >{\centering\arraybackslash}m{0.7cm}    
}
\toprule
\textbf{Methods} & \textbf{2} & \textbf{10} & \textbf{20} & \textbf{50} & \textbf{100} \\
\midrule
DUSt3R             &  3.39  & 42.93 & 142.54 &      &      \\
MASt3R             & 13.16  & 39.17 & 88.61 & 1086.80 &  \\
VGGT               &  1.63  &  \underline{5.42} &  \underline{10.56} &  \underline{29.67} &  \underline{63.06} \\
Fast3R             &  \underline{0.71}  &  \textbf{0.83} &  \textbf{1.72} &   \textbf{5.47} &   \textbf{15.59} \\
MVSAnywhere        &   N/A  &  N/A  & N/A  & 57.47 & 101.51 \\
MVSFormer++        &  14.57   &  34.47  & 55.53     &  121.09 &  268.59 \\
PatchmatchNet      &  4.32   &  9.91  & 21.84     & 121.09 & 268.59 \\
FoundationStereo   & 2.81  &     &      &     &      \\
Stereo4D           & \textbf{0.65}   &     &     &      &      \\
\midrule
COLMAP           &  21.65 &   104.89  & 239.41  &  634.99   &  1358.68   \\

\bottomrule
\end{tabular}
\caption{Runtime (s) of different methods on Urban area across all subsets with varying numbers of input images (2, 10, 20, 50, and 100). “N/A” indicates no result is available.}
\label{tab:runtime_urban}
\end{table}

The processing time was measured separately for each fixed subset of input images (2, 10, 20, 50, and 100 views) selected from the same image block. As shown in Table \ref{tab:runtime_island} and Table \ref{tab:runtime_urban}, the runtime of all methods grows rapidly with the increase in the number of input images. Traditional methods such as COLMAP exhibit extremely high computational cost in the dense reconstruction stage, reaching 1326.38s on 100 images for the Island dataset. Moreover, COLMAP frequently faces issues such as incomplete registered frames and inaccurate feature correspondences, which lead to limited scene coverage and potential reconstruction artifacts.

 Geometry-guided MVS methods are built on top of COLMAP’s sparse reconstruction pipeline, relying on its camera registration and sparse point cloud as input. When COLMAP fails to generate a complete sparse model, the downstream geometry-guided methods cannot proceed with dense reconstruction. As a result, MVSAnywhere, MVSFormer++, and PatchmatchNet fail to produce any results in cases where COLMAP cannot establish the sparse reconstruction. In Table \ref{tab:runtime_island}, when the number of input views is small, the large viewpoint gaps between neighboring images cause significant appearance changes and reduce the overlap area, making correspondence estimation unstable. The island scene contains visually similar textures such as vegetation and rooftops, these repetitive patterns further increase the ambiguity of feature matching under large viewpoint differences. In contrast, with more input views, the accumulated overlap improves the spatial consistency and allows the model to recover a coherent 3D structure.

 In contrast, end-to-end MVS methods only require input images and directly infer scene geometry. Among them, Stereo4D and FoundationStereo are fundamentally stereo-based methods, relying on pairwise image matching. While they achieve fast inference on two-view inputs, they are unable to handle multi-view geometry. The limitation highlights their restricted applicability in real-world aerial photogrammetry scenarios. DUSt3R achieves stable results across different subset sizes, but its computational cost increases rapidly as the number of images grows, which limits its scalability. MASt3R exhibits the highest runtime, reaching over 3 hours for 100 images on each dataset, making it unsuitable for large-scale aerial mapping tasks. In contrast, VGGT and Fast3R demonstrate superior efficiency and scalability, with significantly lower runtime even at large image counts, as summarized in Table \ref{tab:runtime_island} and Table \ref{tab:runtime_urban}.

\begin{table}[]
\centering
\footnotesize
\renewcommand{\arraystretch}{1.0}
\setlength{\tabcolsep}{4pt}
\begin{tabular}{
  >{\centering\arraybackslash}m{0.5cm}
  >{\centering\arraybackslash}m{1.4cm}
  >{\centering\arraybackslash}m{1.5cm}
  >{\centering\arraybackslash}m{1.3cm}
  >{\centering\arraybackslash}m{1.7cm}
}
\toprule
\textbf{Images} & \textbf{Methods} & \textbf{RMS (GSD)} & \textbf{MD (GSD)} & \textbf{Coverage (\%)} \\
\midrule
\multirow{4}{*}{2} 
& DUSt3R & \textbf{2.88} & \textbf{2.21} & \textbf{84.89} \\
& MASt3R & 4.53 & 3.26 & 71.85 \\
& Fast3R & 8.76 & 6.85 & 38.26 \\
& VGGT   & \underline{3.71} & \underline{2.58} & \underline{77.77} \\
\midrule
\multirow{3}{*}{10} 
& DUSt3R & \underline{9.33} & \underline{7.19} & 38.85 \\
& Fast3R & 12.73 & 7.72 & \underline{46.02} \\
& VGGT & \textbf{5.73} & \textbf{3.82} & \textbf{80.65} \\
\midrule
\multirow{2}{*}{20} 
& DUSt3R & \underline{9.70} & \underline{7.38} & \underline{39.88} \\
& VGGT   & \textbf{4.46} & \textbf{2.92} & \textbf{74.32} \\
\bottomrule
\end{tabular}
\caption{Quantitative results of different methods on the Island scene under various numbers of input images. GSD = 0.267m/pixel.}
\label{tab:island-results}
\end{table}

\subsubsection{Accuracy}

\noindent

For large-scale inputs (50 and 100 images), several methods failed to produce valid point clouds or suffered from severe layering artifacts. Therefore, we only report quantitative results for the 2, 10, and 20 image subsets. Fig. \ref{failisland} and Fig. \ref{failurban} illustrate these failed results on the island and urban scenes using different methods. Certain areas are zoomed in to facilitate clearer observation. The reconstructed point clouds exhibit noticeable layering and deformation artifacts, especially over vegetation, rock surfaces, and water regions. These issues mainly arise from unreliable depth estimation and weak geometric constraints in textureless and highly repetitive areas, resulting in distorted 3D structures.

The quantitative accuracy evaluation is reported in Table~\ref{tab:island-results} and Table~\ref{tab:urban-results}, where RMS and MD are used to measure reconstruction errors. The reconstructed point clouds are first coarsely aligned to the reference model, followed by a fine registration using the ICP algorithm in CloudCompare, with the C2C distance threshold set to 5 m.

FoundationStereo and Stereo4D fail to produce usable results on our datasets due to strong distortions and discontinuities in the reconstructed point clouds. Consequently, we exclude them from the quantitative analysis. Among the end-to-end approaches, DUSt3R achieves the best performance in sparse-view conditions. With only 2 input images, it reaches 2.88 GSD RMS and 2.21 MD on the Island scene, and 5.50 GSD RMS and 4.11 GSD MD on the Urban scene. However, as the number of input views increases, DUSt3R’s accuracy drops significantly, particularly in the Island scene, where RMS rises from 2.88 GSD to 9.70 GSD. It indicates reduced geometric consistency and incomplete reconstruction in overlapping regions. VGGT demonstrates superior robustness to increasing view counts, maintaining relatively stable accuracy. MASt3R performs competitively in 2-view cases but struggles to scale. Fast3R generally produces the largest RMS and MD due to insufficient global consistency.

\begin{figure*}[t!]
\begin{center}
    \includegraphics[width=6.5in]{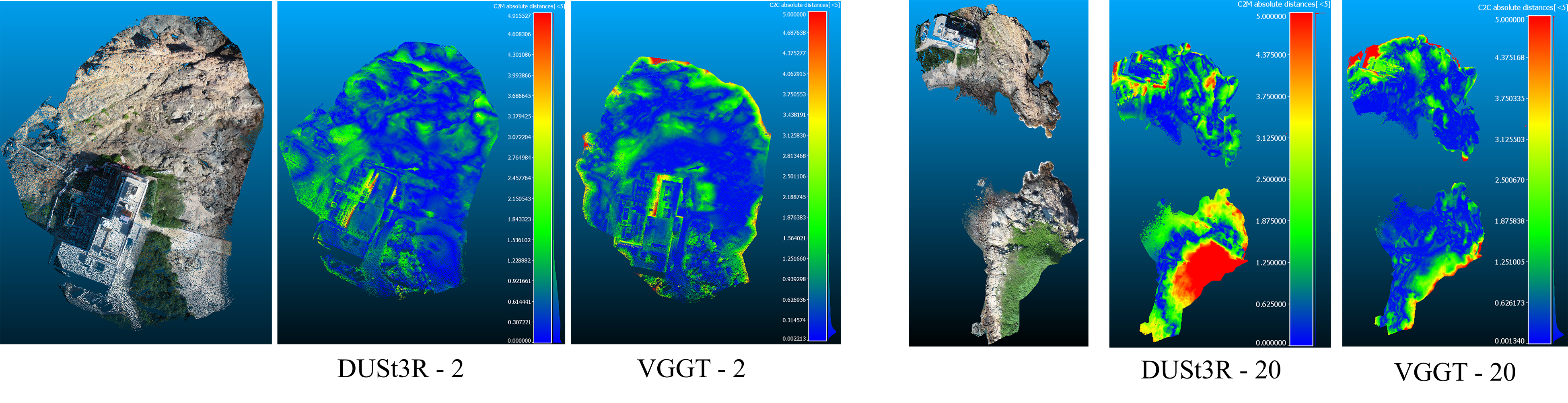}  % Adjusts the figure to the width of the page
    \captionsetup{justification=centering, singlelinecheck=false}
    \centering
    \caption{Comparison of reconstruction accuracy on the Island scene using DUSt3R and VGGT with 2 and 20 input images. The colored maps show point-wise distance errors with respect to the LiDAR ground truth. Blue regions indicate small deviations (higher accuracy), whereas red regions denote larger errors. The color scale represents the error magnitude in meters.}
    \label{c2cisland}
\end{center}
\end{figure*}

\begin{figure*}[t!]
\begin{center}
    \includegraphics[width=6.5in]{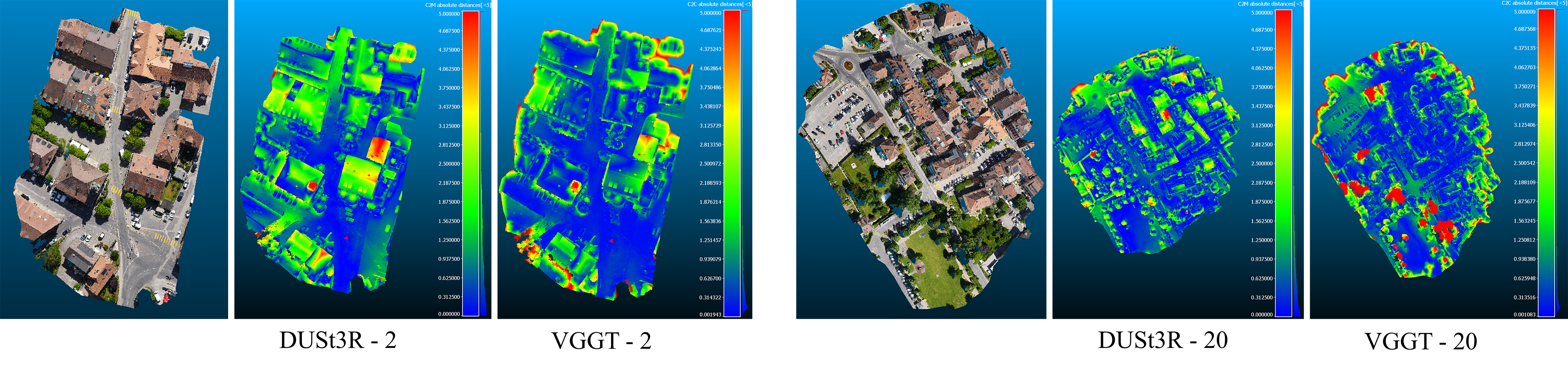}  % Adjusts the figure to the width of the page
    \captionsetup{justification=centering, singlelinecheck=false}
    \centering
    \caption{Comparison of reconstruction accuracy on the Urban scene using DUSt3R and VGGT with 2 and 20 input images. The colored maps show point-wise distance errors with respect to the Pix4dmapper-generated ground truth. Blue regions indicate small deviations (higher accuracy), whereas red regions denote larger errors. The color scale represents the error magnitude in meters.}
    \label{c2curban}
\end{center}
\end{figure*}

These trends are also visually evident in the maps shown in Fig.~\ref{c2cisland} and Fig.~\ref{c2curban}, where reconstruction errors increase significantly in the outer regions of the reconstructed area as the number of input views grows. The errors are particularly pronounced along the scene boundaries and edges, where image overlap is lower and geometric constraints are weaker.

Overall, these results indicate that DUSt3R excels in sparse-view settings, delivering the highest accuracy when image overlap is limited. VGGT shows greater robustness as the number of input views increases. MASt3R and Fast3R exhibit larger RMS and MD values, consistent with the layering and local misalignment artifacts observed in their reconstructed point clouds.

\begin{table}[]
\centering
\footnotesize
\renewcommand{\arraystretch}{1.0}
\setlength{\tabcolsep}{4pt}
\begin{tabular}{
  >{\centering\arraybackslash}m{0.5cm}
  >{\centering\arraybackslash}m{1.4cm}
  >{\centering\arraybackslash}m{1.5cm}
  >{\centering\arraybackslash}m{1.3cm}
  >{\centering\arraybackslash}m{1.7cm}
}
\toprule
\textbf{Images} & \textbf{Methods} & \textbf{RMS (GSD)} & \textbf{MD (GSD)} & \textbf{Coverage (\%)} \\
\midrule
\multirow{4}{*}{2}
& DUSt3R & \underline{5.50} & 4.11 & 55.26 \\
& MASt3R & \textbf{3.87} & \textbf{2.62} & \textbf{78.53} \\
& Fast3R & 6.74 & 5.14 & 45.91 \\
& VGGT   & 5.60 & \underline{3.83} & \underline{65.06} \\
\midrule
\multirow{3}{*}{10}
& DUSt3R & \textbf{3.69} & \textbf{2.55} & \textbf{75.69} \\
& Fast3R & \underline{5.35} & 3.83 & 62.09 \\
& VGGT & 5.46 & \underline{3.69} & \underline{66.71} \\
\midrule
\multirow{4}{*}{20}
& DUSt3R & \textbf{4.33} & \textbf{3.09} & \textbf{68.26} \\
& Fast3R & \underline{5.21} & \underline{3.69} & 63.58 \\
& VGGT  & 5.64 & 3.72 & \underline{66.46} \\
& COLMAP & 8.55 & 6.06 & 52.75 \\
\bottomrule
\end{tabular}
\caption{Quantitative results of different methods on the Urban scene under various numbers of input images. GSD = 0.282 m/pixel}
\label{tab:urban-results}
\end{table}

\subsubsection{Coverage}

\noindent

The coverage values reported in Table~\ref{tab:island-results} and Table~\ref{tab:urban-results} provide a quantitative measure of the proportion of ground truth points effectively reconstructed by different methods.  

DUSt3R achieves the highest coverage in sparse-view settings, with coverage rates of 84.89\% in the Island scene and 55.26\% in the Urban scene using only 2 input images. However, its coverage collapses when more inputs are introduced, particularly on the Island scene.

In contrast, VGGT maintains the most stable and balanced coverage across different input sizes and scenes. Its reconstruction completeness decreases moderately with increasing views. When the number of input images increases, other methods exhibit less favorable scaling behavior in terms of reconstruction accuracy and stability. Fast3R maintains low coverage in general. MASt3R performs well only in sparse-view settings but degrades with more inputs.

\section{Conclusions and Future Work}

This paper presents a comparative study of traditional, geometry-guided, and end-to-end learning-based MVS methods for aerial photogrammetry, evaluated on island and urban scenes. The results show clear differences among the three paradigms. Traditional methods, represented by COLMAP, provide relatively sparse reconstruction results under controlled conditions. These also suffer from high computational cost and poor scalability in large-scale outdoor environments. Geometry-guided methods inherit this dependency on upstream sparse reconstruction, leading to frequent failures when image registration is incomplete. In contrast, end-to-end approaches demonstrate higher robustness and better generalization, with VGGT achieving the best balance between reconstruction quality and runtime efficiency. In contrast, end-to-end approaches provide faster reconstruction and better coverage. However, they still produce large reconstruction errors, limiting their practical usability. Common issues, such as layering artifacts and local structural inconsistencies, still limit their practical deployment in high-precision aerial mapping.

In the future, we will focus on integrating pose-aware priors such as SLAM trajectories to enhance large-scale reconstruction accuracy, developing scalable inference strategies to process large image collections efficiently.

	\begin{spacing}{1.17}
		\normalsize
		\bibliography{ISPRSguidelines_authors} % Include your own bibliography (*.bib), style is given in isprs.cls
	\end{spacing}

\end{document}